\documentclass{article}
\usepackage{spconf,amsmath,epsfig}
\begin{document}\sloppy
\def\x{{\mathbf x}}
\def\L{{\cal L}}

\title{Dynamic Texture Synthesis By Incorporating Long-range Spatial 
and Temporal Correlations}

\name{Kaitai Zhang, Bin Wang, Hong-Shuo Chen, Ye Wang, Shiyu Mou, and C.-C. Jay Kuo
}
\address{University of Southern California, Los Angeles, USA}

\maketitle

\begin{abstract}
The main challenge of dynamic texture synthesis lies in how to maintain
spatial and temporal consistency in synthesized videos. The
major drawback of existing dynamic texture synthesis models comes
from poor treatment of the long-range texture correlation and motion
information. To address this problem, we incorporate a new loss term,
called the Shifted Gram loss, to capture the structural and long-range
correlation of the reference texture video. Furthermore, we introduce a
frame sampling strategy to exploit long-period motion across multiple
frames. With these two new techniques, the application scope of existing
texture synthesis models can be extended. That is, they can
synthesize not only homogeneous but also structured dynamic texture
patterns. Thorough experimental results are provided to demonstrate
that our proposed dynamic texture synthesis model offers
state-of-the-art visual performance. 
\end{abstract}


\section{Introduction}\label{sec:intro}

Given a short video clip of target dynamic texture as the reference,
the dynamic texture synthesis task is to synthesize dynamic texture
video of arbitrary length. Understanding, characterizing, and
synthesizing temporal texture patterns has been a problem of interest in
human perception, computer vision, and pattern recognition in recent
years. Examples of such video patterns span from amorphous matter like
flame, smoke, water to well-shaped items like waving flags, flickering
candles, and living fountains. A large amount of work has been done on
dynamic texture patterns \cite{heeger1986seeing, nelson1992qualitative,
doretto2003dynamic, xie2018learning, han2019learning}, including dynamic
texture classification, and dynamic texture synthesis.  Research on
texture video sequences finds numerous real-world applications, including
fire detection, foreground and background separation, and generic video
analysis. Furthermore, characterizing these temporal patterns have
theoretical significance in understanding the mechanism behind human
perception of temporal correlation of video data. 

As compared with static texture image synthesis, the dynamic texture
synthesis problem lies in the 3D space. That is, it needs not only to
generate an individual frame as a static texture image but also to
process the temporal information to build a coherent image sequence.
The main challenge in dynamic texture study is to model the motion
behavior (or dynamics) of texture elements. It is a non-trivial and
challenging problem.  Thanks to the rapid development and superior
performance of deep learning methods, many papers have been published
with amazing visual effects in dynamic texture synthesis. Despite the
progress, there are still several drawbacks in existing models as 
pointed out in \cite{tesfaldet2018two} and \cite{funke2017synthesising}. 

\begin{figure}[tb]
\begin{minipage}[b]{0.15\linewidth}
  \centering
  \centerline{\includegraphics[width=1.3cm]{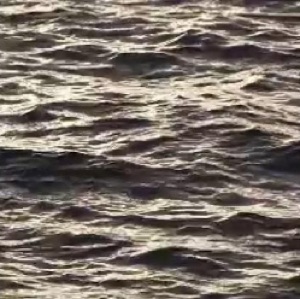}}
  \medskip
\end{minipage}
\hfill
\begin{minipage}[b]{0.15\linewidth}
  \centering
  \centerline{\includegraphics[width=1.3cm]{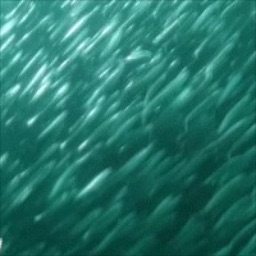}}
  \medskip
\end{minipage}
\hfill
\begin{minipage}[b]{0.15\linewidth}
  \centering
  \centerline{\includegraphics[width=1.3cm]{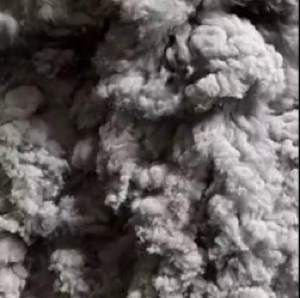}}
  \medskip
\end{minipage}
\hfill
\begin{minipage}[b]{0.15\linewidth}
  \centering
  \centerline{\includegraphics[width=1.3cm]{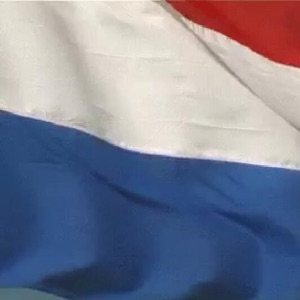}}
  \medskip
\end{minipage}
\hfill
\begin{minipage}[b]{0.15\linewidth}
  \centering
  \centerline{\includegraphics[width=1.3cm]{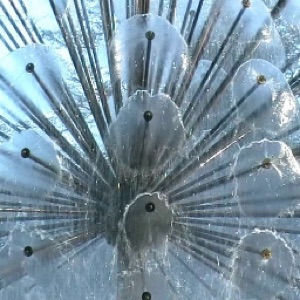}}
  \medskip
\end{minipage}
\hfill
\begin{minipage}[b]{0.15\linewidth}
  \centering
  \centerline{\includegraphics[width=1.3cm]{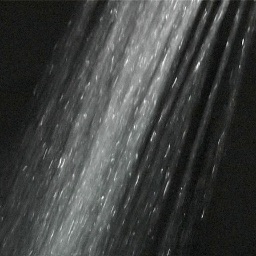}}
  \medskip
\end{minipage}

\begin{minipage}[b]{0.15\linewidth}
  \centering
  \centerline{\includegraphics[width=1.3cm]{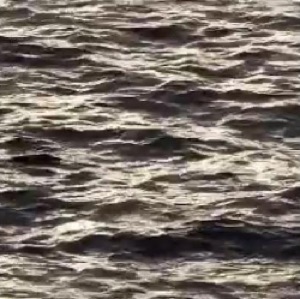}}
  \medskip
\end{minipage}
\hfill
\begin{minipage}[b]{0.15\linewidth}
  \centering
  \centerline{\includegraphics[width=1.3cm]{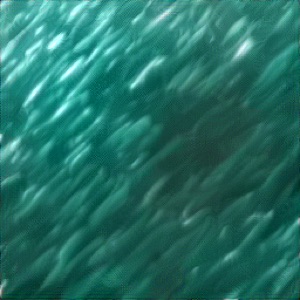}}
  \medskip
\end{minipage}
\hfill
\begin{minipage}[b]{0.15\linewidth}
  \centering
  \centerline{\includegraphics[width=1.3cm]{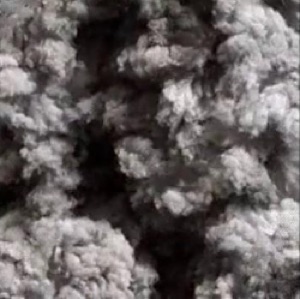}}
  \medskip
\end{minipage}
\hfill
\begin{minipage}[b]{0.15\linewidth}
  \centering
  \centerline{\includegraphics[width=1.3cm]{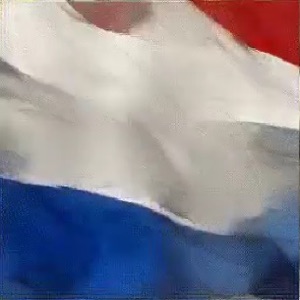}}
  \medskip
\end{minipage}
\hfill
\begin{minipage}[b]{0.15\linewidth}
  \centering
  \centerline{\includegraphics[width=1.3cm]{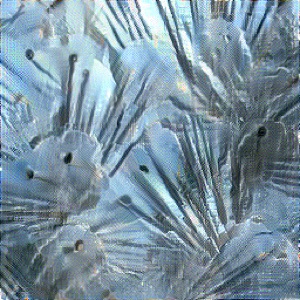}}
  \medskip
\end{minipage}
\hfill
\begin{minipage}[b]{0.15\linewidth}
  \centering
  \centerline{\includegraphics[width=1.3cm]{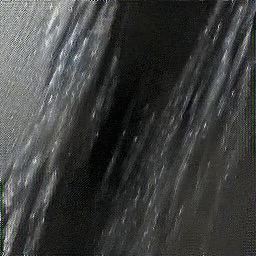}}
  \medskip
\end{minipage}

\begin{minipage}[b]{0.15\linewidth}
  \centering
  \centerline{\includegraphics[width=1.3cm]{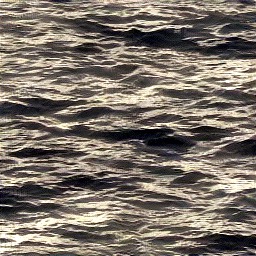}}
  \medskip
\end{minipage}
\hfill
\begin{minipage}[b]{0.15\linewidth}
  \centering
  \centerline{\includegraphics[width=1.3cm]{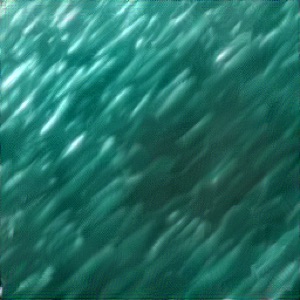}}
  \medskip
\end{minipage}
\hfill
\begin{minipage}[b]{0.15\linewidth}
  \centering
  \centerline{\includegraphics[width=1.3cm]{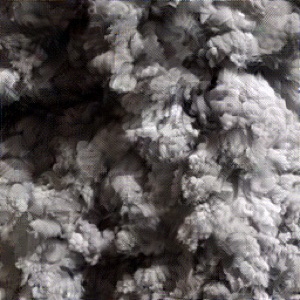}}
  \medskip
\end{minipage}
\hfill
\begin{minipage}[b]{0.15\linewidth}
  \centering
  \centerline{\includegraphics[width=1.3cm]{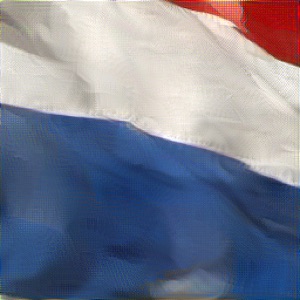}}
  \medskip
\end{minipage}
\hfill
\begin{minipage}[b]{0.15\linewidth}
  \centering
  \centerline{\includegraphics[width=1.3cm]{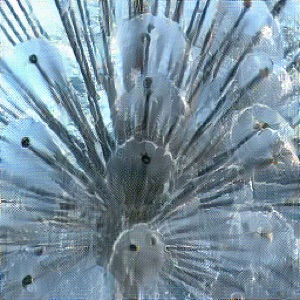}}
  \medskip
\end{minipage}
\hfill
\begin{minipage}[b]{0.15\linewidth}
  \centering
  \centerline{\includegraphics[width=1.3cm]{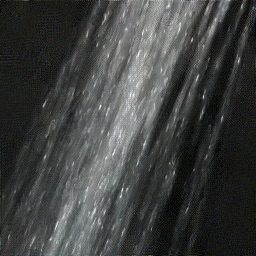}}
  \medskip
\end{minipage}
\caption{This figure gives six representative dynamic texture examples
(one column per texture), where the left three are homogeneous textures
while the right three are structured textures.  The first row shows one
frame of reference dynamic texture video, the second row shows
synthesized results obtained by the baseline model, and the third row
shows results obtained using our model.} \label{fig:1}
\end{figure}

The first limitation is the existing model fails for texture examples
that have the long-range spatial correlation. Some examples are shown in
the 4th, 5th and 6th columns of Fig. \ref{fig:1}. As discussed in
previous work \cite{tesfaldet2018two,funke2017synthesising}, this
drawback is attributed to the fact that the current loss function (i.e.,
the Gram loss) is not efficient in characterizing the long-range spatial
correlation. It is shown by experiments that the Gram loss can provide
excellent performance in capturing local image properties such as
texture patterns. Yet, it cannot handle long-range image information
well since such information is discarded in the modeling process.
Actually, we observe that existing solutions fail to provide
satisfactory performance in synthesizing dynamic textures with
long-range as well as mid-range correlations. 

The second drawback is that the current model can generate textures with
monotonous motion, i.e. relatively smooth motion between adjacent
frames.  In other words, the generated samples do not have diversified
dynamics. Sometimes, they even appear to be periodic. This is because
that previous models primarily focus on dynamics between adjacent
frames, but ignore motion in a longer period.  For example, Funke {\em
et al.} \cite{funke2017synthesising} used the correlation matrix between
frames to represent the motion information, and Tesfaldet {\em et al.}
\cite{tesfaldet2018two} adopted a network branch for optical flow
prediction to learn dynamics. This shortcoming yields visible distortion
to the perceptual quality of synthesized results.

Being motivated by the above two drawbacks, we propose a new solution to
address them in this paper. First, we incorporate a new loss term,
called the Shifted Gram loss, to capture the structural and long-range
correlation of the reference texture video. Second, we introduce a frame
sampling strategy to exploit long-period motion across multiple frames.
The solution is implemented using an enhanced two-branch convolutional
neural network.  It can synthesize dynamic texture with long-range
spatial and temporal correlations. As shown in Fig. \ref{fig:1}, the
proposed method could take care of both homogeneous and structured
texture patterns well. Extensive experimental results will be given to
show the superiority of the proposed method. 


\section{Related Work}\label{sec:review}

Texture analysis on classification and segmentation has been studied for decades and great success have been made in the past several years \cite{zhang2019texture,zhang2019data}. 
Recently, more research attention has been paid to static texture image synthesis problem,
{\em e.g.,}
\cite{berger2016incorporating, zhou2018non,sendik2017deep}.  Berger and
Memisevic \cite{berger2016incorporating} used a variation of the Gram
loss that takes spatial co-occurrences of local features into account.
Zhou {\em et al.} \cite{zhou2018non} proposed a generative adversarial
network (GAN) that was trained to double the spatial extent of texture
blocks extracted from a specific texture exemplar.  Sendik and Cohen-Or
\cite{sendik2017deep} introduced a structural energy loss term that
captures self-similar and regular characteristics of textures based on
correlations among deep features. 

However, in spite of these efforts, dynamic texture synthesis remains to
be a challenging and non-trivial problem for several reasons. First,
even it is possible to capture and maintain the structural information
in a single image, it is still difficult to preserve the structural
information in an image sequence. Second, there exist more diversified
patterns in dynamic textures due to various spatio-temporal arrangement.
More analysis with illustrative examples will be elaborated in Sec.
\ref{sec:Exp}. To summarize, extending a static texture synthesis model
to a dynamic one is not a straightfoward problem, which will be the focus
of our current work.

\section{Proposed Method}\label{sec:method}

The baseline model to be used in our work is the two-stream CNN proposed
by Tesfaldet {\em et al.} \cite{tesfaldet2018two}. In this section, we
will briefly introduce the baseline model, including the network
structure, the learning method and the synthesis process. Then, we will
propose several techniques to address long-range spatial and temporal
correlations in the underlying reference texture sequence. 

\subsection{Baseline Model}\label{subsec:baseline}

The baseline model is constructed from two convolutional networks
(CNNs), an appearance stream and a dynamics stream. Such kind of
two-stream design enables the model to factorize appearance and dynamics
of dynamic texture and proceed with the analysis independently. Similar
to previous work on textures synthesis, the networks summarizes an input
dynamic texture in terms of a set of activation statistics of filter
outputs,and then build a handbook for each texture of interest in a
dictionary learning manner. 

During the synthesis process, it optimizes a randomly initialized noise
pattern such that its spatio-temporal statistics from each stream match
those of the input texture. To achieve that, the model conducts Gradient
Descent and Back-propagation during the synthesis process, and it
optimizes the loss function with respect to each pixel to match the
spatio-temporal statistics. Meanwhile, all the parameters from object
recognition and optical flow prediction networks are pre-trained and
remain fixed during the synthesis process. 

\textbf{Appearance Stream.} The design of appearance stream follows the
spatial texture model which is first introduced by
\cite{gatys2015texture}. We use the same publicly available normalized
VGG19 network\cite{simonyan2014very} which is first used by
\cite{gatys2015texture}. Previous research has shown that such kind of
CNN pre-trained on an object classification task can be very effective
at characterizing texture appearance. 

To describe the appearance of an input dynamic texture, we first feed
each frame of input video into the network, and denote the feature maps
at layer $l$ as $F^{l}_{i} \in M_{l}$, where $F^{l}_{i}$ is the $i^{th}$
vectorized feature map of layer $l$, and $M_{l}$ is the number of
entrances in each feature map of layer $l$. Then, the pair-wise product
of each vectorized feature map, i.e. Gram matrix is defined as
\begin{equation}
G^{l}_{ij} = \frac{1}{M_{l}}\sum^{M_{l}}_{k}{F^{l}_{ik}F^{l}_{jk}} = 
\frac{1}{M_{l}}F^{l}_{i}\cdot F^{l}_{j},
\end{equation}
where $\cdot$ means inner product. Following previous works, we use
feature maps at the following five layers: Conv layer 1, Pooling layer
1, Pooling layer 2, Pooling layer 3, Pooling layer 4. 

In the synthesis process, we initialize a random noise sequence
$\hat{I}$, and then feed them into network to proceed with its Gram
matrix representation $\hat{G}$. Then the apparent loss could be defined as
\begin{equation}
\mathcal{L}^{l}_{Appearance} = \frac{1}{M_{l}}\sum_{i,j}(G^{l}_{ij} 
- \hat{G}^{l}_{ij})^{2}
\end{equation}
The final apparent loss is weight sum from all selected layers defined as 
\begin{equation}
\mathcal{L}_{Appearance} = \sum_{l}w_{l}\mathcal{L}^{l}_{Appearance}.
\end{equation}

\textbf{Dynamics stream.} The dynamic stream follows the baseline
model \cite{tesfaldet2018two} employing a pre-trained optical flow
prediction network, which takes each pair of temporally consecutive
frames as input. The structure of dynamic stream is described as
following. The first layer consists of 32 $3D$ convolution filters of
spatial size $11\times11$. Then, a squaring activation function and
$5\times5$ spatial max-pooling with a stride of one are applied
sequentially. Then, a $1\times1$ convolution layer follows with 64
filters follows. Finally, to remove local contrast dependence, a L1
divisive normalization is applied. To capture dynamic, i.e. texture
motion, in multiple scales, a image pyramid is employed, and image at
each scale is processed independently and concatenated to get the final
feature. 

To describe the dynamic of an input dynamic texture, we feed each pair
of consecutive frames into the network, and conduct exactly same Gram
matrix computing as the appearance stream. Here, we use $D$ to denote
Gram matrix in dynamic stream. During the synthesis process, we could
have $\hat{D}$ computed from the generated dynamic texture $\hat{I}$.
Then the dynamic loss could be defined as
\begin{equation}
\mathcal{L}^{l}_{Dynamic} = 
\frac{1}{M_{l}}\sum_{i,j}(D^{l}_{ij} - \hat{D}^{l}_{ij})^{2}.
\end{equation}
The final apparent loss is weighted sum from all selected layers defined as 
\begin{equation}
\mathcal{L}_{Dynamic} = \sum_{l}w_{l}\mathcal{L}^{l}_{Dynamic}.
\end{equation}

\begin{figure}[tb]
\begin{minipage}[b]{0.49\linewidth}
  \centering
  \centerline{\includegraphics[width=4cm]{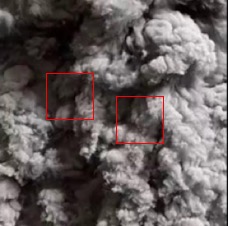}}
  \medskip
\end{minipage}
\hfill
\begin{minipage}[b]{0.49\linewidth}
  \centering
  \centerline{\includegraphics[width=4cm]{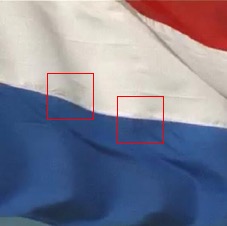}}
  \medskip
\end{minipage}
\caption{This figure shows the importance of local feature correlation
in structural texture. The red boxes above show a pair of texture
patches in the same location. And those patches could be viewed as
visualization of the receptive field in a certain layer.} \label{fig:2}
\end{figure}

\subsection{Proposed Method - I}\label{subsec:method-1}

The effectiveness of Gram matrices could be explained as it captures the
coherence across multiple feature maps at a single location. In another
word, if we view the pre-trained CNN as a well-designed filter sets, the
correlation between responses of different filter sets describe the
appearance of texture. And this kind of coherence is so powerful for
texture that we could even use a shallow network with random weights to
synthesize static texture, even without the help from the complicated
pre-trained CNN. And the similar methods have been adopted in many other
computer vision topics, like image style transfer and impainting.
However, as shown in \ref{sec:intro}, it still has non-negligible
drawbacks. 

Form the equation, we could find that the Gram matrices are totally
blind to the global arrangement of objects inside each frame of the
reference video. Specifically, we could see that it makes pair-wise
feature production in each single location and then take the spatial
averaging. As a result, all spatial information is erased during the
inner product calculation, which makes it fail to capture structural
information in a relative large scale. Herein, inspired by
\cite{berger2016incorporating}, we integrate the following shifted Gram
matrices to replace the original Gram matrices. 
\begin{equation}\label{Eq:T2}
\hat{G}^{l}_{ij} = \frac{1}{M_{l}} T(F^{l}_{i})\cdot T(F^{l}_{j}),
\end{equation}
where $T$ is a spatial transform operator. By this way, we could further
capture the correlation between local feature and feature from its
neighborhood. And we could adapt various spatial transform with
different angel and aptitude. Gram matrices only make use of
correlations between feature maps whereas image structures are
represented by the correlations inside each feature map. 

The importance and effectiveness of those spatial correlation between
local feature and its neighborhood could be illustrated with
\ref{fig:2}. If the dynamic texture has a homogeneous appearance, the
two texture patches will have very weak correlation. For example, in
homogeneous texture like smoke, if we fix the left patch in the
synthesis process, we could have many possible texture appearance in the
location of right patch, as long as it consists with its local neighbor
in a natural way. On the contrary, in a structural texture like flag,
those two patches in the same location will have a stronger dependency.
If we fix one of them, the other one must obey the constraint. 

Then, the appearance loss with a spatial transform operator $T$ could be
computed accordingly. And we could designed a combination of different
$T$ to get a comprehensive appearance loss. 

\subsection{Proposed Method - II}\label{subsec:method-2}

\begin{figure}[tb]
\begin{minipage}[b]{0.95\linewidth}
  \centering
  \centerline{\includegraphics[width=8cm]{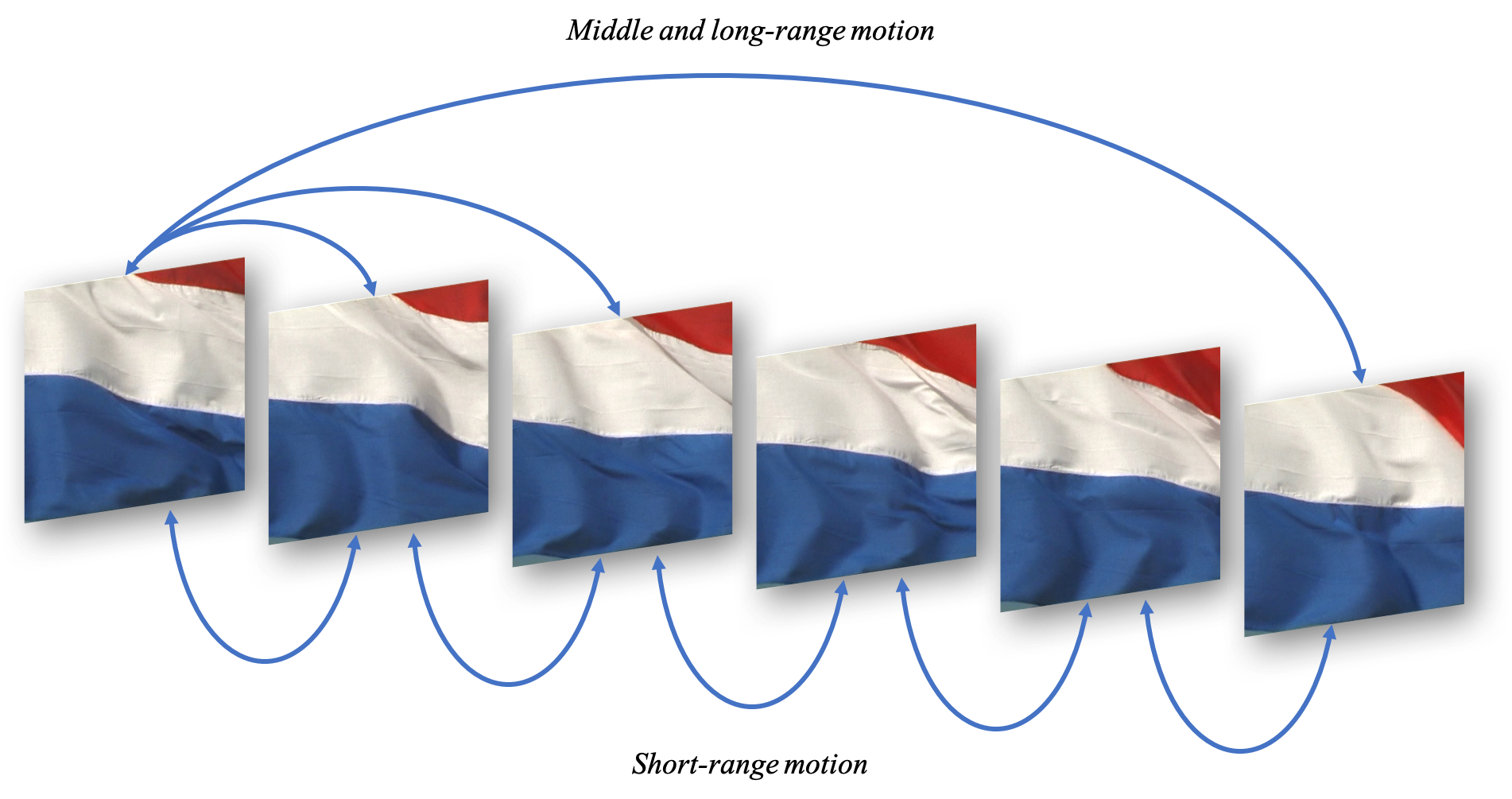}}
  \medskip
\end{minipage}
\caption{This figure shows the importance of middle and long-range
motion, Using flag sequence as an example, we can find that most
meaningful motion are actually crossing multiple frames. Such kind of
long-range motion combines with long-range spatial structure give us a
vivid visual effect together. }
\end{figure}

There two critical requirements to synthesize perceptually natural
textures: spatial consistency and temporal consistency. The proposed
method in the previous section is designed to render textures when it
has long-range structure within each frame. In other words, it handles
the non-uniformity in spatial domain. So, a very straightforward
extension is that, is there existing non-uniformity in temporal domain?
And can the previous model handle that? More specifically, here we use
non-uniformity in temporal domain to refer the long-time and non-local
motion in dynamic texture. 

Actually, this is a quite obvious problem when we try to generalize the
model from homogeneous texture to structure texture. First, texture with
long-range structure naturally owns more complicated motion patterns
with time elapsing. Second, maintaining long-range structures stable and
continuous in continuous frames will be much harder comparing with
homogeneous texture. Oppositely, the synthesized result from the
previous model often looks periodic, which could be a huge drawback and
damage the perceptual quality of those synthesized results. Moreover, we
also observe the previous model could generate some local motion hopping
in continuous frames, which also implies the temporal consistency need
to be improved. 

In the baseline model, it uses an optical flow prediction network to get
features that capture the dynamic between every two consecutive frames.
And those features are further utilized in the synthesis process. So, in
the previous method, all temporal dynamic in synthesis results is from
frame pairs between two consecutive frames, and those temporal dynamic
features are further averaged over the whole video. That makes the model
only learn the temporal motion/dynamic in a very short time period, and
ignore any potential long-range temporal pattern in the reference video. 

Here, we employ a simple yet effective approach to encode both
short-range motion and long-range motion. We proposed a multi-period
frame sampling strategy to modify the dynamic loss. Rather than simply
using each consecutive frame pairs, we first set a sampling interval
$t$, and then take $i$th frame with $i+t$th frame as a pair to compute
the dynamic loss with interval $t$ as follow
\begin{equation}
\mathcal{L}_{Dynamic}(t) = 
\sum_{i,l}\beta_{l}\mathcal{L}^{l}_{Dynamic}(t;i).
\end{equation}

The final dynamic loss is the weighted sum of all selected time intervals
\begin{equation}
\mathcal{L}_{Dynamic} = \sum_{t}\alpha_{t}\mathcal{L}^{l}_{Dynamic}(t).
\end{equation}

\section{Experiment}\label{sec:Exp}

In this section, we show our experimental results on both homogeneous
and structural dynamic texture video, and then further compare the
proposed method with the baseline model. Given their temporal nature,
our results are best viewed as videos.

\begin{figure}[tb]
\begin{minipage}[b]{0.15\linewidth}
  \centering
  \centerline{\includegraphics[width=1.6cm]{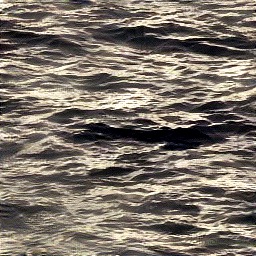}}
  \medskip
\end{minipage}
\hfill
\begin{minipage}[b]{0.15\linewidth}
  \centering
  \centerline{\includegraphics[width=1.6cm]{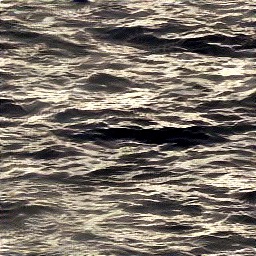}}
  \medskip
\end{minipage}
\hfill
\begin{minipage}[b]{0.15\linewidth}
  \centering
  \centerline{\includegraphics[width=1.6cm]{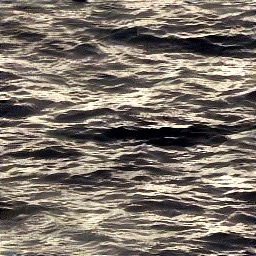}}
  \medskip
\end{minipage}
\hfill
\begin{minipage}[b]{0.15\linewidth}
  \centering
  \centerline{\includegraphics[width=1.6cm]{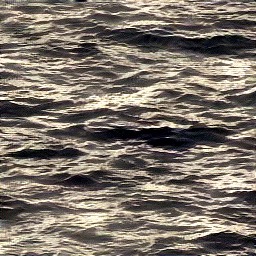}}
  \medskip
\end{minipage}
\hfill
\begin{minipage}[b]{0.15\linewidth}
  \centering
  \centerline{\includegraphics[width=1.6cm]{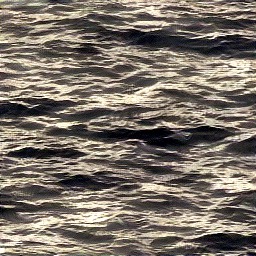}}
  \medskip
\end{minipage}

\begin{minipage}[b]{0.15\linewidth}
  \centering
  \centerline{\includegraphics[width=1.6cm]{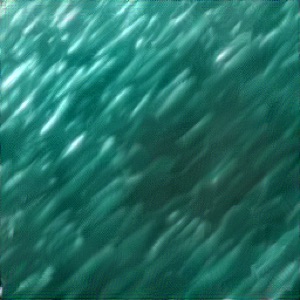}}
  \medskip
\end{minipage}
\hfill
\begin{minipage}[b]{0.15\linewidth}
  \centering
  \centerline{\includegraphics[width=1.6cm]{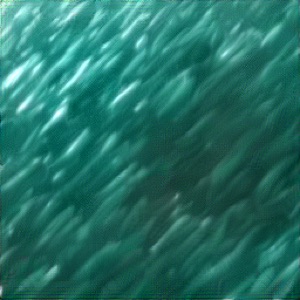}}
  \medskip
\end{minipage}
\hfill
\begin{minipage}[b]{0.15\linewidth}
  \centering
  \centerline{\includegraphics[width=1.6cm]{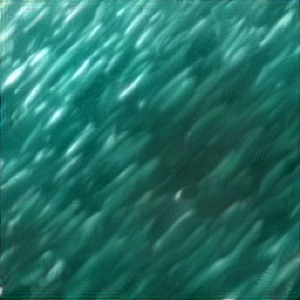}}
  \medskip
\end{minipage}
\hfill
\begin{minipage}[b]{0.15\linewidth}
  \centering
  \centerline{\includegraphics[width=1.6cm]{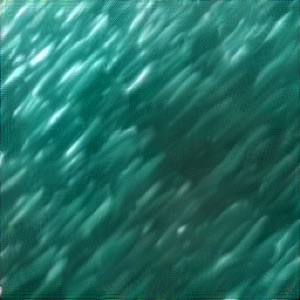}}
  \medskip
\end{minipage}
\hfill
\begin{minipage}[b]{0.15\linewidth}
  \centering
  \centerline{\includegraphics[width=1.6cm]{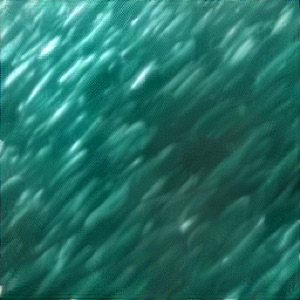}}
  \medskip
\end{minipage}

\begin{minipage}[b]{0.15\linewidth}
  \centering
  \centerline{\includegraphics[width=1.6cm]{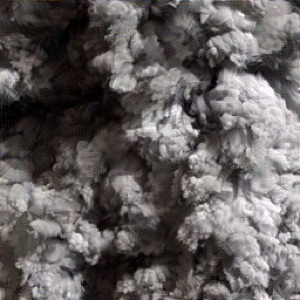}}
  \medskip
\end{minipage}
\hfill
\begin{minipage}[b]{0.15\linewidth}
  \centering
  \centerline{\includegraphics[width=1.6cm]{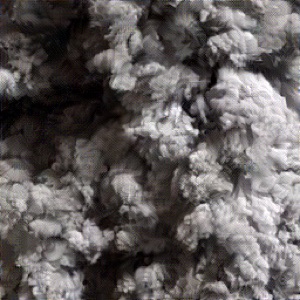}}
  \medskip
\end{minipage}
\hfill
\begin{minipage}[b]{0.15\linewidth}
  \centering
  \centerline{\includegraphics[width=1.6cm]{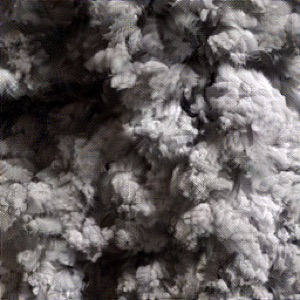}}
  \medskip
\end{minipage}
\hfill
\begin{minipage}[b]{0.15\linewidth}
  \centering
  \centerline{\includegraphics[width=1.6cm]{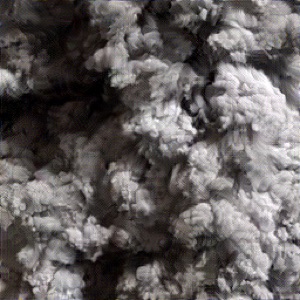}}
  \medskip
\end{minipage}
\hfill
\begin{minipage}[b]{0.15\linewidth}
  \centering
  \centerline{\includegraphics[width=1.6cm]{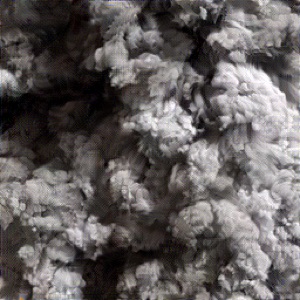}}
  \medskip
\end{minipage}

\caption{This figure give several examples of homogeneous dynamic texture frame by frame. }
\end{figure}

\subsection{Experiment Setting}\label{subsec:subhead}

Our two-stream architecture is implemented using TensorFlow. Results
were generated using an NVIDIA Titan X GPU and synthesis time ranges
between 10 to 30 minutes to generate 8 frames with an image resolution
of 256 by 256. To achieve a fair comparison, all dynamic texture samples
shown in this paper are selected from DynTex
Dataset \cite{peteri2010dyntex} following the baseline
model \cite{tesfaldet2018two}. 

Dynamic textures are implicitly defined as the local minimum of this
loss function. Textures are generated by optimizing with respect to the
pixels of the video. Diversified results could be obtained by
initializing the optimization process using I.I.D. Gaussian noise, and
the non-convex property of the loss function could also provide extra
variations. Consistent with previous
works \cite{gatys2015texture,tesfaldet2018two}, we use
L-BFGS \cite{liu1989limited} optimizer. 

For appearance stream, we use Conv layer 1, Pooling layer 1, Pooling
layer 2, Pooling layer 3, Pooling layer 4 to compute
$\mathcal{L}_{Appearance}$. For proposed technique I, here we use
horizontal and vertical transform as our $T$, and the shift distances
are set as 8, 16, 32, and 128. For proposed technique II, we choose time
interval as 1, 2, and 4. All other settings remain the same as baseline
to achieve a fair comparison. 

\begin{figure}[tb]
\label{fig:4}
\begin{minipage}[b]{0.15\linewidth}
  \centering
  \centerline{\includegraphics[width=1.6cm]{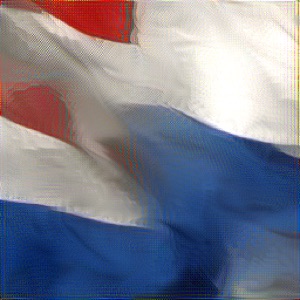}}
  \medskip
\end{minipage}
\hfill
\begin{minipage}[b]{0.15\linewidth}
  \centering
  \centerline{\includegraphics[width=1.6cm]{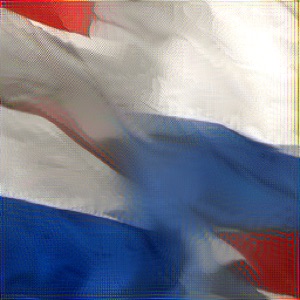}}
  \medskip
\end{minipage}
\hfill
\begin{minipage}[b]{0.15\linewidth}
  \centering
  \centerline{\includegraphics[width=1.6cm]{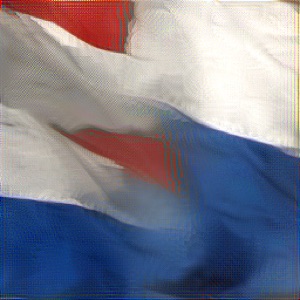}}
  \medskip
\end{minipage}
\hfill
\begin{minipage}[b]{0.15\linewidth}
  \centering
  \centerline{\includegraphics[width=1.6cm]{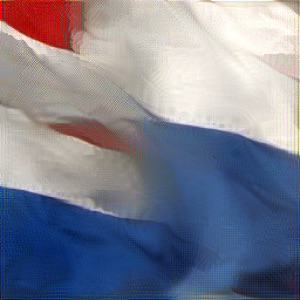}}
  \medskip
\end{minipage}
\hfill
\begin{minipage}[b]{0.15\linewidth}
  \centering
  \centerline{\includegraphics[width=1.6cm]{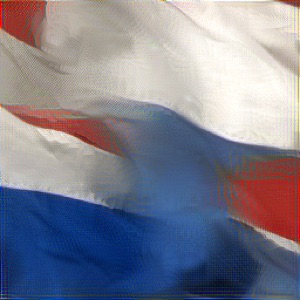}}
  \medskip
\end{minipage}

\begin{minipage}[b]{0.15\linewidth}
  \centering
  \centerline{\includegraphics[width=1.6cm]{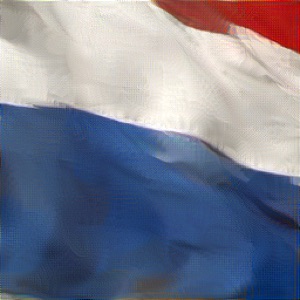}}
  \medskip
\end{minipage}
\hfill
\begin{minipage}[b]{0.15\linewidth}
  \centering
  \centerline{\includegraphics[width=1.6cm]{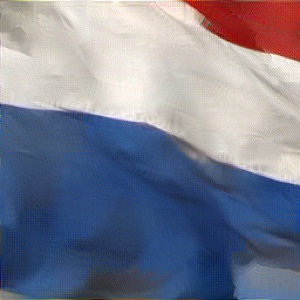}}
  \medskip
\end{minipage}
\hfill
\begin{minipage}[b]{0.15\linewidth}
  \centering
  \centerline{\includegraphics[width=1.6cm]{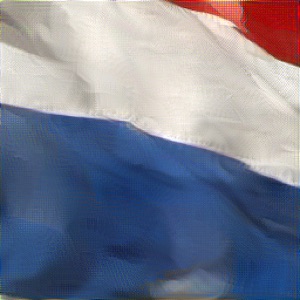}}
  \medskip
\end{minipage}
\hfill
\begin{minipage}[b]{0.15\linewidth}
  \centering
  \centerline{\includegraphics[width=1.6cm]{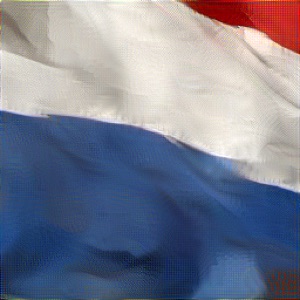}}
  \medskip
\end{minipage}
\hfill
\begin{minipage}[b]{0.15\linewidth}
  \centering
  \centerline{\includegraphics[width=1.6cm]{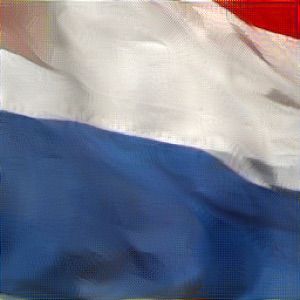}}
  \medskip
\end{minipage}
  \medskip

\begin{minipage}[b]{0.15\linewidth}
  \centering
  \centerline{\includegraphics[width=1.6cm]{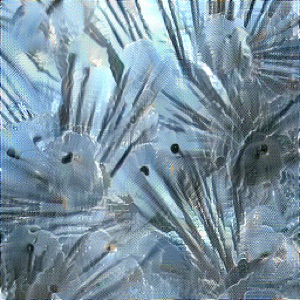}}
  \medskip
\end{minipage}
\hfill
\begin{minipage}[b]{0.15\linewidth}
  \centering
  \centerline{\includegraphics[width=1.6cm]{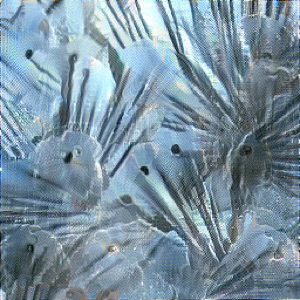}}
  \medskip
\end{minipage}
\hfill
\begin{minipage}[b]{0.15\linewidth}
  \centering
  \centerline{\includegraphics[width=1.6cm]{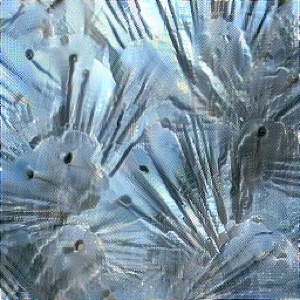}}
  \medskip
\end{minipage}
\hfill
\begin{minipage}[b]{0.15\linewidth}
  \centering
  \centerline{\includegraphics[width=1.6cm]{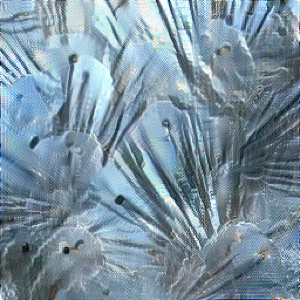}}
  \medskip
\end{minipage}
\hfill
\begin{minipage}[b]{0.15\linewidth}
  \centering
  \centerline{\includegraphics[width=1.6cm]{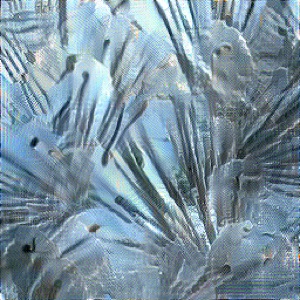}}
  \medskip
\end{minipage}

\begin{minipage}[b]{0.15\linewidth}
  \centering
  \centerline{\includegraphics[width=1.6cm]{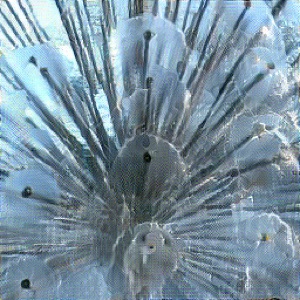}}
  \medskip
\end{minipage}
\hfill
\begin{minipage}[b]{0.15\linewidth}
  \centering
  \centerline{\includegraphics[width=1.6cm]{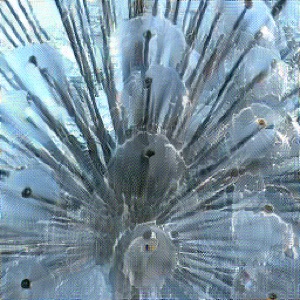}}
  \medskip
\end{minipage}
\hfill
\begin{minipage}[b]{0.15\linewidth}
  \centering
  \centerline{\includegraphics[width=1.6cm]{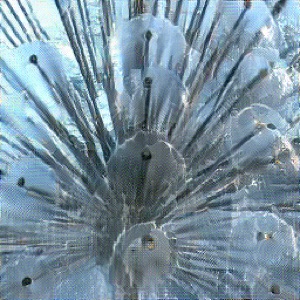}}
  \medskip
\end{minipage}
\hfill
\begin{minipage}[b]{0.15\linewidth}
  \centering
  \centerline{\includegraphics[width=1.6cm]{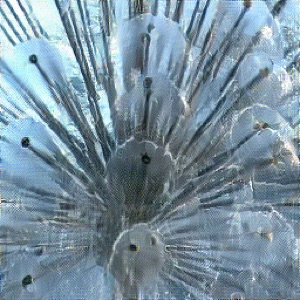}}
  \medskip
\end{minipage}
\hfill
\begin{minipage}[b]{0.15\linewidth}
  \centering
  \centerline{\includegraphics[width=1.6cm]{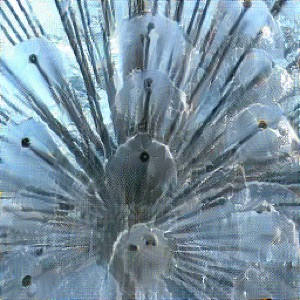}}
  \medskip
\end{minipage}
  \medskip

\begin{minipage}[b]{0.15\linewidth}
  \centering
  \centerline{\includegraphics[width=1.6cm]{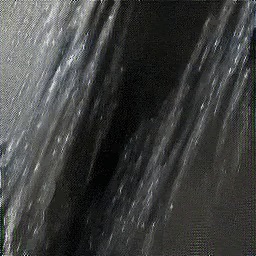}}
  \medskip
\end{minipage}
\hfill
\begin{minipage}[b]{0.15\linewidth}
  \centering
  \centerline{\includegraphics[width=1.6cm]{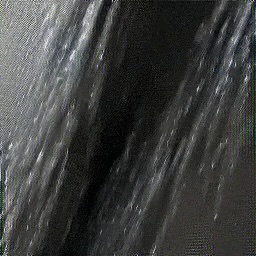}}
  \medskip
\end{minipage}
\hfill
\begin{minipage}[b]{0.15\linewidth}
  \centering
  \centerline{\includegraphics[width=1.6cm]{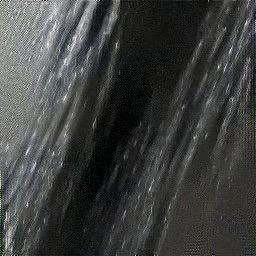}}
  \medskip
\end{minipage}
\hfill
\begin{minipage}[b]{0.15\linewidth}
  \centering
  \centerline{\includegraphics[width=1.6cm]{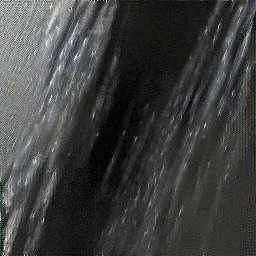}}
  \medskip
\end{minipage}
\hfill
\begin{minipage}[b]{0.15\linewidth}
  \centering
  \centerline{\includegraphics[width=1.6cm]{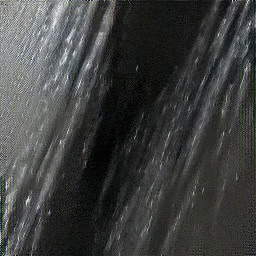}}
  \medskip
\end{minipage}

\begin{minipage}[b]{0.15\linewidth}
  \centering
  \centerline{\includegraphics[width=1.6cm]{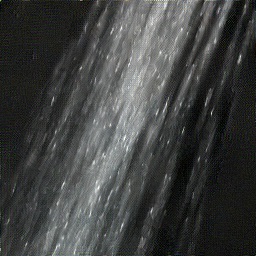}}
  \medskip
\end{minipage}
\hfill
\begin{minipage}[b]{0.15\linewidth}
  \centering
  \centerline{\includegraphics[width=1.6cm]{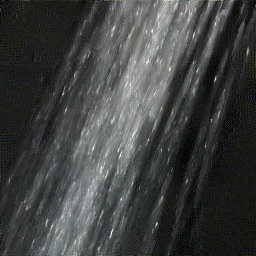}}
  \medskip
\end{minipage}
\hfill
\begin{minipage}[b]{0.15\linewidth}
  \centering
  \centerline{\includegraphics[width=1.6cm]{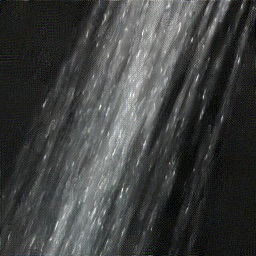}}
  \medskip
\end{minipage}
\hfill
\begin{minipage}[b]{0.15\linewidth}
  \centering
  \centerline{\includegraphics[width=1.6cm]{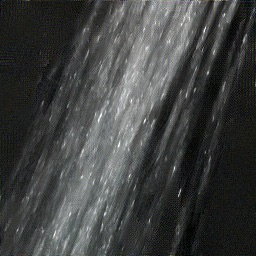}}
  \medskip
\end{minipage}
\hfill
\begin{minipage}[b]{0.15\linewidth}
  \centering
  \centerline{\includegraphics[width=1.6cm]{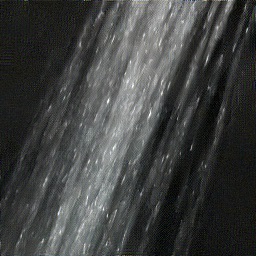}}
  \medskip
\end{minipage}
  \medskip
\caption{This figure gives several examples of dynamic textures which
contain structural information with long-range correlations. }
\end{figure}

\subsection{Experimental Results}\label{subsec:results}

We further conduct experiments on several typical dynamic texture
sequence which contain strong structural information. To achieve a fair
comparison, here we list synthesized results from both baseline and our
model, frame by frame. 

\textbf{Homogeneous Dynamic Texture.} We first test our methods on
several typical homogeneous dynamic texture sequences. The purpose is to
verify that the proposed method could maintain the advantages of
baseline model and keep the ability of synthesizing excellent
homogeneous dynamic textures. Fig.4 shows some examples of synthesized
dynamic texture frame by frame. 

\begin{figure}[tb]
\begin{minipage}[b]{0.15\linewidth}
  \centering
  \centerline{\includegraphics[width=1.6cm]{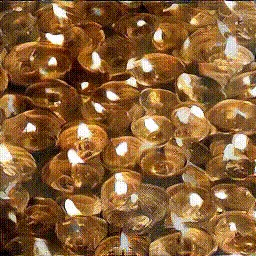}}
  \medskip
\end{minipage}
\hfill
\begin{minipage}[b]{0.15\linewidth}
  \centering
  \centerline{\includegraphics[width=1.6cm]{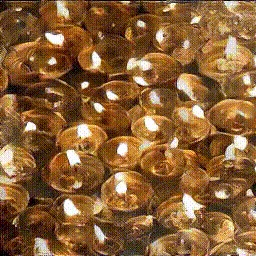}}
  \medskip
\end{minipage}
\hfill
\begin{minipage}[b]{0.15\linewidth}
  \centering
  \centerline{\includegraphics[width=1.6cm]{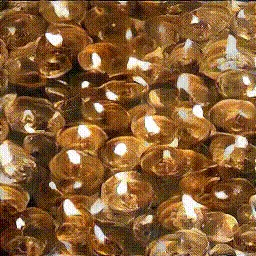}}
  \medskip
\end{minipage}
\hfill
\begin{minipage}[b]{0.15\linewidth}
  \centering
  \centerline{\includegraphics[width=1.6cm]{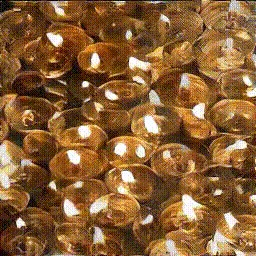}}
  \medskip
\end{minipage}
\hfill
\begin{minipage}[b]{0.15\linewidth}
  \centering
  \centerline{\includegraphics[width=1.6cm]{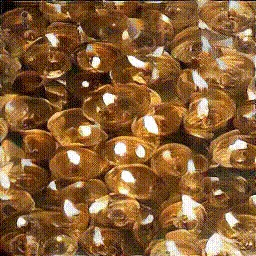}}
  \medskip
\end{minipage}

\begin{minipage}[b]{0.15\linewidth}
  \centering
  \centerline{\includegraphics[width=1.6cm]{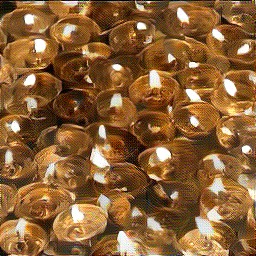}}
  \medskip
\end{minipage}
\hfill
\begin{minipage}[b]{0.15\linewidth}
  \centering
  \centerline{\includegraphics[width=1.6cm]{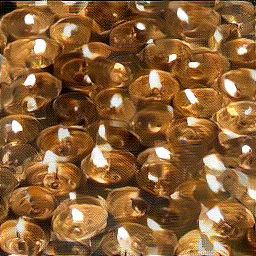}}
  \medskip
\end{minipage}
\hfill
\begin{minipage}[b]{0.15\linewidth}
  \centering
  \centerline{\includegraphics[width=1.6cm]{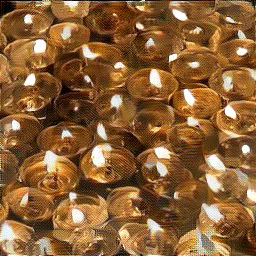}}
  \medskip
\end{minipage}
\hfill
\begin{minipage}[b]{0.15\linewidth}
  \centering
  \centerline{\includegraphics[width=1.6cm]{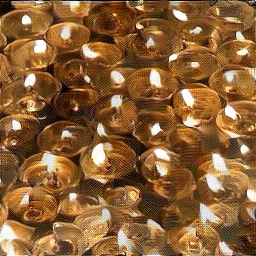}}
  \medskip
\end{minipage}
\hfill
\begin{minipage}[b]{0.15\linewidth}
  \centering
  \centerline{\includegraphics[width=1.6cm]{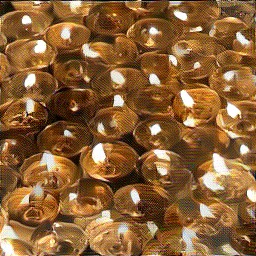}}
  \medskip
\end{minipage}
  \medskip

\begin{minipage}[b]{0.15\linewidth}
  \centering
  \centerline{\includegraphics[width=1.6cm]{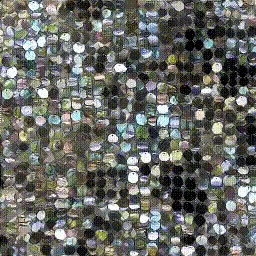}}
  \medskip
\end{minipage}
\hfill
\begin{minipage}[b]{0.15\linewidth}
  \centering
  \centerline{\includegraphics[width=1.6cm]{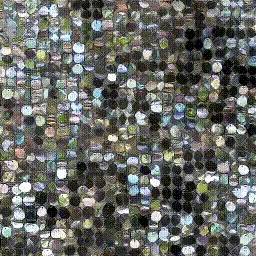}}
  \medskip
\end{minipage}
\hfill
\begin{minipage}[b]{0.15\linewidth}
  \centering
  \centerline{\includegraphics[width=1.6cm]{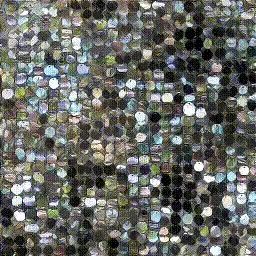}}
  \medskip
\end{minipage}
\hfill
\begin{minipage}[b]{0.15\linewidth}
  \centering
  \centerline{\includegraphics[width=1.6cm]{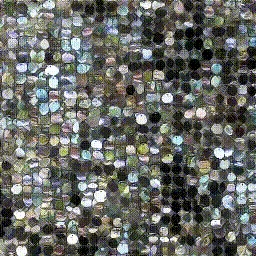}}
  \medskip
\end{minipage}
\hfill
\begin{minipage}[b]{0.15\linewidth}
  \centering
  \centerline{\includegraphics[width=1.6cm]{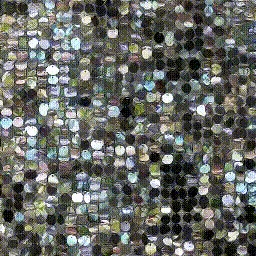}}
  \medskip
\end{minipage}

\begin{minipage}[b]{0.15\linewidth}
  \centering
  \centerline{\includegraphics[width=1.6cm]{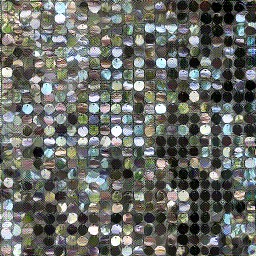}}
  \medskip
\end{minipage}
\hfill
\begin{minipage}[b]{0.15\linewidth}
  \centering
  \centerline{\includegraphics[width=1.6cm]{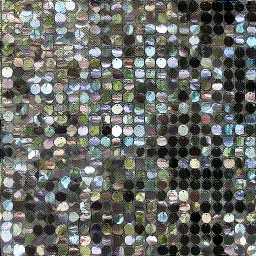}}
  \medskip
\end{minipage}
\hfill
\begin{minipage}[b]{0.15\linewidth}
  \centering
  \centerline{\includegraphics[width=1.6cm]{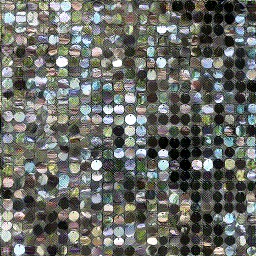}}
  \medskip
\end{minipage}
\hfill
\begin{minipage}[b]{0.15\linewidth}
  \centering
  \centerline{\includegraphics[width=1.6cm]{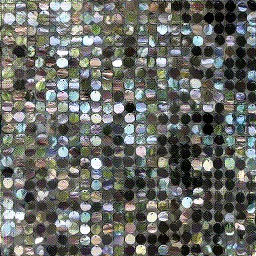}}
  \medskip
\end{minipage}
\hfill
\begin{minipage}[b]{0.15\linewidth}
  \centering
  \centerline{\includegraphics[width=1.6cm]{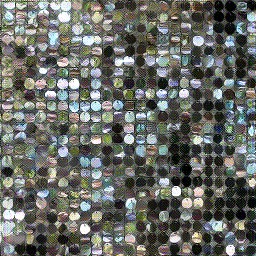}}
  \medskip
\end{minipage}
  \medskip

\begin{minipage}[b]{0.15\linewidth}
  \centering
  \centerline{\includegraphics[width=1.6cm]{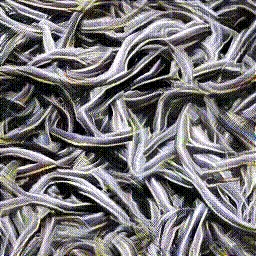}}
  \medskip
\end{minipage}
\hfill
\begin{minipage}[b]{0.15\linewidth}
  \centering
  \centerline{\includegraphics[width=1.6cm]{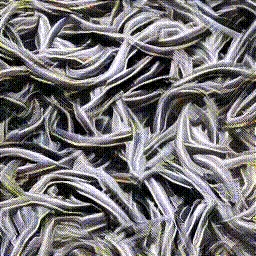}}
  \medskip
\end{minipage}
\hfill
\begin{minipage}[b]{0.15\linewidth}
  \centering
  \centerline{\includegraphics[width=1.6cm]{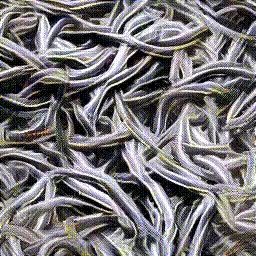}}
  \medskip
\end{minipage}
\hfill
\begin{minipage}[b]{0.15\linewidth}
  \centering
  \centerline{\includegraphics[width=1.6cm]{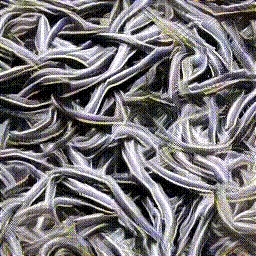}}
  \medskip
\end{minipage}
\hfill
\begin{minipage}[b]{0.15\linewidth}
  \centering
  \centerline{\includegraphics[width=1.6cm]{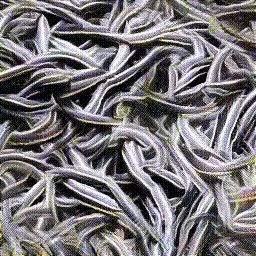}}
  \medskip
\end{minipage}

\begin{minipage}[b]{0.15\linewidth}
  \centering
  \centerline{\includegraphics[width=1.6cm]{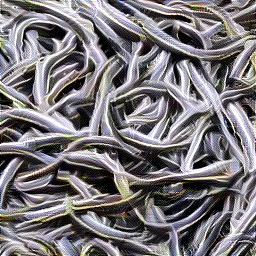}}
  \medskip
\end{minipage}
\hfill
\begin{minipage}[b]{0.15\linewidth}
  \centering
  \centerline{\includegraphics[width=1.6cm]{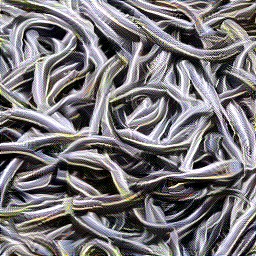}}
  \medskip
\end{minipage}
\hfill
\begin{minipage}[b]{0.15\linewidth}
  \centering
  \centerline{\includegraphics[width=1.6cm]{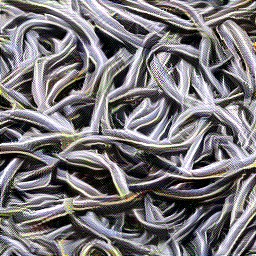}}
  \medskip
\end{minipage}
\hfill
\begin{minipage}[b]{0.15\linewidth}
  \centering
  \centerline{\includegraphics[width=1.6cm]{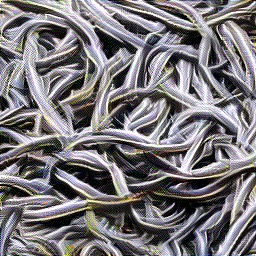}}
  \medskip
\end{minipage}
\hfill
\begin{minipage}[b]{0.15\linewidth}
  \centering
  \centerline{\includegraphics[width=1.6cm]{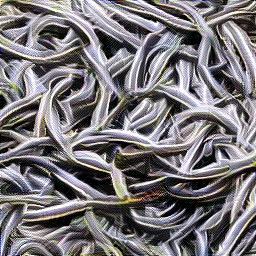}}
  \medskip
\end{minipage}
  \medskip
\caption{This figure gives several examples of dynamic texture which
contain structural information with middle-range correlations.}
\end{figure}

\textbf{Structure with long-range correlation.} As for dynamic texture
with long-range structure, our proposed method outperforms the baseline
model on all these sequences, and several typical exampled are shown in
Fig. 5. Noteworthily, proposed techniques work well even all those
sequences have quite different forms of long-range correlation. In the
flag sequences, the structural information externalizes as the boundary
between two colors, which is an irregular curve and across the whole
dimension. To make a natural synthesis, the boundary must be sharp and
clear in each frame, and also has wavy motion as the other part of the
flag. In the fountain sequence, the structural information is shown by
the stable fountain skeleton behind the flowing water. Unlike the flag,
the fountain sequence requires those edge stable, and any motion will
lead to failure results. In the shower sequence, the difficulty is how
to teach the model to learn the discrete water trajectory as well as the
uniform background. As more results shown on our project page, there are
more kinds of different form of long-range correlations in dynamic
texture sequences, but our proposed method could handle them in an
unified manner. 

\textbf{Structure with middle-range correlation.} We also notice that
there are a lot of texture in between of homogeneous texture and
structural texture. As shown in Fig.6, some kind of dynamic textures
have structures which is not global and obvious, but such kind of
structures are still critical for perceived quality of our synthesized
results. For example, in the candle sequence, we must keep edge of those
candles approximated circular. Similarly, we need to keep the snake's
torso natural and smooth in the snake sequence. 

Another interesting thing about such kind of sequences, as we point out
in previous section, is the results from baseline model show many
unexpected sudden local hopping at local structure in consecutive
frames. Due to their temporal nature, we recommend to view those results
in video form to fully understand this phenomenon.The reason behind
those flaws, as we analyzed, is the lack of enough constraint for those
local structures leads to discontinuity in the time dimension. By
introducing better activation statistics, our model shows better results
on such kind of sequences as well. 

We have explored proposed techniques thoroughly and found a few
limitations which we leave as potential future improvements. First,
although dynamic textures are highly self-similar in the spatial domain,
the temporal motion in a dynamic texture video is much more
complicated(Fig. 2). Second, like most previous models and some works in
similar areas, the generated video only has a relatively low resolution.
In other words, those generated videos are more like visual effects but
not real videos with vivid details. So, there is still a non-neglected
distance to generate dynamic textures with better temporal motion in a
higher resolution. 

\section{Conclusion}\label{sec:conclusion}

Two effective techniques for dynamic texture synthesis were presented
and proved effective. Compared with the baseline model, the enhanced
model could encode coherence of local features as well as the
correlation between local feature and its neighbors, and also capture
more complicated motion in the time domain. It was shown by extensive
experimental results that the proposed method offers state-of-the-art
performance. 

\bibliographystyle{IEEEbib}
\bibliography{icme2020template}

\end{document}